\documentclass[11pt]{article}

\usepackage[preprint]{acl}

\usepackage{times}
\usepackage{latexsym}

\usepackage[T1]{fontenc}

\usepackage[utf8]{inputenc}

\usepackage{microtype}

\usepackage{inconsolata}

\usepackage{graphicx}
\usepackage{makecell}
\usepackage{booktabs}
\usepackage{enumitem}
\usepackage{amsmath}
\usepackage{amsfonts}

\definecolor{DarkGreen}{RGB}{0,120,0}

\title{Evolutionary Search for Automated Design \\of Uncertainty Quantification Methods}

\author{
 \textbf{Mikhail Seleznyov\textsuperscript{1,2}},
 \textbf{Daniil Korbut\textsuperscript{3}},
 \textbf{Viktor Moskvoretskii\textsuperscript{4}},
\\
 \textbf{Oleg Somov\textsuperscript{1,5}},
 \textbf{Alexander Panchenko\textsuperscript{1,2}},
 \textbf{Elena Tutubalina\textsuperscript{1}}
\medskip\\
 \textsuperscript{1}AIRI,
 \textsuperscript{2}Skoltech,
 \textsuperscript{3}Amazon,\\
 \textsuperscript{4}EPFL,
 \textsuperscript{5}MIPT
\\
}

\begin{document}
\maketitle
\begin{abstract}
Uncertainty quantification (UQ) methods for large language models are predominantly designed by hand based on domain knowledge and heuristics, limiting their scalability and generality. We apply LLM-powered evolutionary search to automatically discover unsupervised UQ methods represented as Python programs. On the task of atomic claim verification, our evolved methods outperform strong manually-designed baselines, achieving up to 6.7\% relative ROC-AUC improvement across 9 datasets while generalizing robustly out-of-distribution. Qualitative analysis reveals that different LLMs employ qualitatively distinct evolutionary strategies: Claude models consistently design high-feature-count linear estimators, while Gpt-oss-120B gravitates toward simpler and more interpretable positional weighting schemes. Surprisingly, only Sonnet 4.5 and Opus 4.5 reliably leverage increased method complexity to improve performance~-- Opus 4.6 shows an unexpected regression relative to its predecessor. Overall, our results indicate that LLM-powered evolutionary search is a promising paradigm for automated, interpretable hallucination detector design.
\end{abstract}

\section{Introduction}

Large Language Models (LLMs) are widely deployed as chatbots and coding assistants, however their propensity to hallucinate is a significant blocker on the path toward reliable and trustworthy general-purpose systems \citep{huang2025asurveyonhallucinationinlargelanguagemodels}. 

An important step in mitigating factual hallucinations is developing robust detectors which work across multiple domains and don't require access to ground truth facts~\cite{vazhentsev2026leveragingllmparametricknowledge}.
A common approach to designing such detectors is based on the uncertainty quantification (UQ) task. The core assumption is that there is a way to quantify an intrinsic quantity -- LLM uncertainty -- such that the estimated uncertainty value is highly correlated with hallucination occurrences.

UQ methods for LLMs typically rely on output distributions or internal representations, and achieve strong empirical performance~\citep{fadeeva2023lmpolygraph}. However, most of these methods are manually designed by researchers, who use domain knowledge and intuition to craft often complex scoring rules, algorithms, or heuristics. Unfortunately, such hand-engineered approaches may be brittle and limited in scalability, and are often viewed as less promising than more automated alternatives~\citep{sutton2019bitterlesson}.


\begin{figure}
    \centering
    \includegraphics[width=\linewidth]{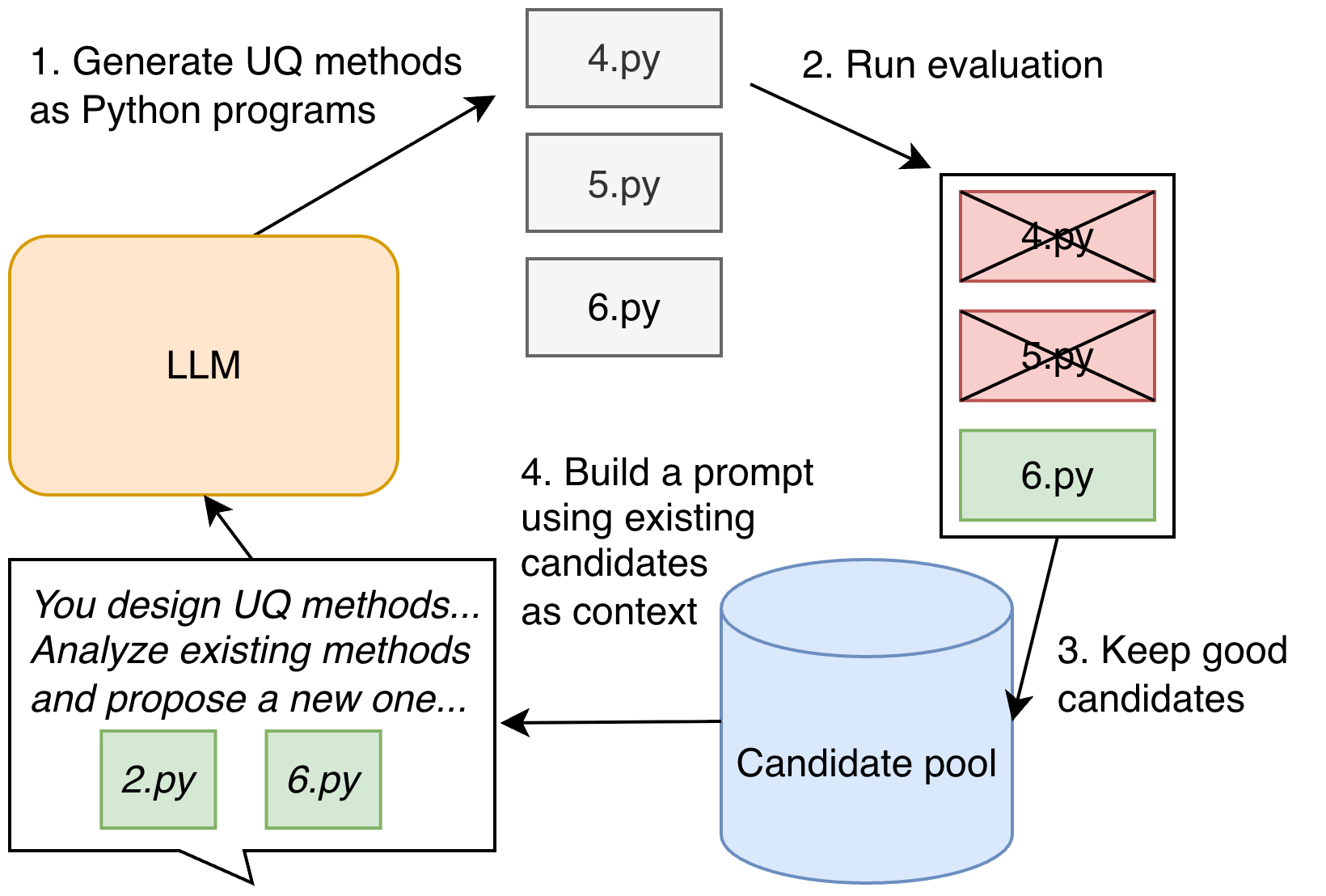}
    \caption{A visualization of LLM-powered evolutionary search pipeline, used to autonomously design uncertainty quantification (UQ) methods for hallucination detection tasks. At the start the candidate pool is initialized with a single selected baseline method.}
    \label{fig:evo_pipeline}
\end{figure}

Meanwhile, recent progress in LLMs has enabled a new direction for scalable automation of complex tasks. AlphaEvolve \citep{novikov2025alphaevolve}, in particular, highlights the potential of LLM-powered evolutionary optimization as a general framework for automated discovery. Such methods have already demonstrated progress on challenging problems, including data-center scheduling, efficient matrix multiplication subroutines, hardware circuit design, and other domains~\citep{openevolve,lange2025shinkaevolve,lee2025evolvingdeeperllmthinking}.


Unlike manual design or end-to-end deep learning, LLM-powered evolutionary search manages to combine two goals that have traditionally been hard to achieve at the same time. First, it offers interpretability, because one can carefully design the search space in advance. Second, it offers scalability, by taking advantage of the ever-improving capabilities of language models. 

Our contributions are along two axes: we show that evolutionary search works for automated design of UQ methods, and we use it as a lens to reveal qualitative differences in how frontier LLMs approach open-ended optimization.
\begin{enumerate}[leftmargin=8pt, itemsep=0.3pt, topsep=0.3pt]
\item \textbf{We propose applying LLM-based evolutionary search to automatically design UQ methods for hallucination detection}. This yields explicit, human-readable scoring functions that combine the scalability of evolutionary optimization with the interpretability of hand-crafted methods.
\item \textbf{We demonstrate that evolved methods surpass strong manually-crafted baselines on atomic claim verification}, achieving statistically significant gains on 6 out of 9 datasets while generalizing robustly out-of-distribution from a single training dataset. Cross-task transfer to selective classification is limited, but in-domain evolution matches existing alternatives without requiring attention maps.
\item \textbf{We show that different frontier LLMs exhibit qualitatively distinct search strategies}, and only certain models reliably leverage increased method complexity to improve performance.
\end{enumerate}

Overall, these results establish LLM-powered evolutionary search as a viable paradigm for scalable, interpretable hallucination detector design, and surface some behavioral differences between frontier models that warrant further investigation.



\section{Related work}

\paragraph{Uncertainty Quantification.}
Uncertainty of LLM predictions is generally a strong signal of answer correctness. Most uncertainty estimators operate on the model output probability distribution, including methods such as Sequence Probability, Mean Token Entropy, and Perplexity~\cite{fadeeva2023lmpolygraph}. Despite their simplicity, these approaches are well aligned with the probabilistic nature of language models and remain difficult to outperform, particularly in cross-domain and cross-task transfer settings~\cite{moskvoretskii2025adaptive}.

More advanced methods leverage internal model signals, such as attention maps (e.g., RAUQ~\cite{rauq}, AttentionScore~\cite{NEURIPS2024_LLM}) or hidden representations (e.g., SATMD and SATRMD~\cite{vazhentsev2025token}). Alternatively, uncertainty can be estimated through \emph{verbalized confidence}, where the LLM explicitly reports its own confidence via prompting~\cite{kadavath2022languagemodelsmostlyknowwhatheyknow}.
All the above methods are manually designed. Some generalize well, but there is no silver bullet, and developing task-specific UQ approaches still relies heavily on researchers' manual effort.

\paragraph{LLM-powered evolutionary search.} Recent work explores using LLMs as mutation and refinement operators inside evolutionary search loops. AlphaEvolve \cite{novikov2025alphaevolve} is a prominent demonstration of this paradigm for scientific and algorithmic discovery, combining LLM-proposed code edits with automated evaluators in an iterative evolutionary pipeline. Follow-up work has focused on improving openness, efficiency, and adaptability of this framework. ShinkaEvolve \cite{lange2025shinkaevolve} improves sample efficiency through better parent selection, novelty-based rejection, and adaptive model selection, while CodeEvolve \cite{assumpo2026codeevolveopensourceevolutionary} and GigaEvo \cite{khrulkov2025gigaevoopensourceoptimization} provide open-source evolutionary coding frameworks aimed at reproducibility and broader adoption. Beyond static search, \citet{surina2025algorithmdiscoverywithllms} propose combining evolutionary search with reinforcement learning so that the LLM itself improves during search, and ThetaEvolve \cite{wang2025thetaevolvetesttimelearningopen} studies test-time learning within evolving open-problem solvers. Related ideas also appear beyond code optimization: Mind Evolution \cite{lee2025evolvingdeeperllmthinking} applies evolutionary generation and recombination to inference-time reasoning, showing that evolutionary search can be a more general mechanism for improving LLM outputs.

\paragraph{Program synthesis.}
Several papers explore application of evolution-like pipelines for generation of programs. \citet{liu2025alphago}
search for novel LLM architectures using a multi-agent approach. \citet{wei2025astra} optimize CUDA kernels for common SGLang \citep{zheng2024sglang} routines, and \citet{press2025algotune} try to speed up general computational routines (e.g. \texttt{np.qr} or \texttt{python.gzip}). However, to the best of our knowledge, there are no works which considered autonomous design of hallucination detectors.

\section{Method}
We focus on developing unsupervised and computationally efficient uncertainty quantification (UQ) methods, which use take in internal states of a small pretrained transformer model, e.g. Llama-3.1-8B-Instruct \citep{llama2024thellama3}, and produce a single real-valued sequence-level uncertainty estimate.

We use EvoTune framework introduced by \citet{surina2025algorithmdiscoverywithllms} since its Docker-based implementation ensures good portability.
The process of evolutionary search is schematically described in Figure \ref{fig:evo_pipeline}. The framework accepts four inputs: a baseline UQ method as a starting point, a dataset $D$ for candidate evaluation, the evolution prompt, and the evolution-driving LLM. A dataset must contain $(x, y)$ pairs, where $x$ is a set of precomputed internal states for a given model generation (logits, residual hidden states, attention maps, or a subset of these), and $y \in \mathbb{R}$ is ground truth quality score of this generation (e.g. True/False for binary classification,  ROUGE-L for summarization, etc). 

\paragraph{Evolution prompt.} The evolution prompt describes the task, inputs available for the UQ methods,  provides design guidelines and includes example candidates from the candidate pool along with their performance on the dataset $D$. At initialization, the candidate pool only contains the baseline method. Then, the sampling procedure depends on the round index. First, top $p_i$ percent of candidates are preselected. Second, a probability distribution $p$ over candidates is computed, with 
\begin{equation*}
    p(\texttt{candidate}) \propto \exp\left(\frac{{\texttt{candidate\_score}}}{T_{\texttt{cand\_sampling}}}\right),
\end{equation*}
where \texttt{candidate\_score} is the target metric of this candidate on the dataset $D$ and $T_{\texttt{cand\_sampling}}$ controls how concentrated the distribution is. Note this is distinct from temperature $T$ used for the evolution LLM.

\paragraph{Main evolution loop.} At each round, several \textbf{candidate} methods are generated by the evolution LLM in form of Python programs, using the same prompt and non-zero temperature $T$. Second, target metrics are computed on the dataset $D$. Third, candidates are stored in the candidate pool. Finally, the \textbf{evolution prompt} is built, based on task-dependent instructions and 1-4 randomly sampled candidates. 
For most \textbf{evolution runs}, we set the number of evolution rounds to 500. We only sample 2 candidates per round, since found that models frequently generate conceptually identical candidates, even at temperature 1.0, which led to inefficient exploration. 
Unless otherwise specified, we use Sequence Probability \cite{fomicheva-etal-2020-unsupervised} as a starting point. 
For the LLM model governing the generation of novel candidates, we consider Gpt-oss-120b \citep{openai2025gptoss120bgptoss20bmodel} as a strong open-source model, as well as a number of closed-source models from Claude family: Sonnet 4.0, Haiku 4.5, Sonnet 4.5, Opus 4.5, Opus 4.6 \citep{anthropic2025claude40,anthropic2025claude45haiku,anthropic2025claude45sonnet,anthropic2025claude45opus,anthropic2025claude46opus}.

\begin{figure*}
    \centering
    \includegraphics[width=\linewidth]{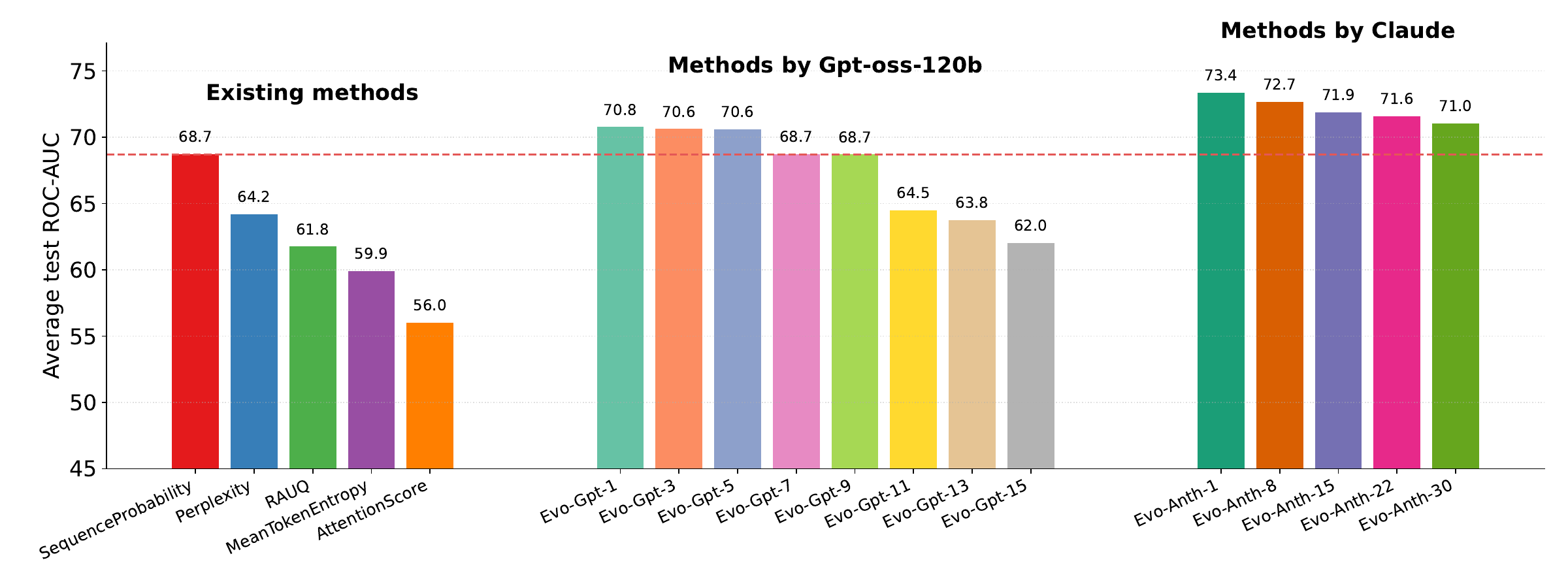}
    \caption{Average ROC-AUC across 9 hallucination detection datasets (atomic factual claims). For Claude-generated candidates, top-30 methods were selected by validation performance on PopQA dataset. We report 5 of them with test-performance ranks 1, 8, 15, 22, and 30 respectively. For Gpt-oss-generated methods, we show every second method among top-16 by validation performance on PopQA. }
    \label{fig:barplot}
\end{figure*}

\section{Benchmarks}

We apply LLM-powered evolutionary search to two tasks, related to hallucination mitigation: atomic claim verification and selective prediction.

\paragraph{Atomic claim verification.}
We evaluate atomic claim verification as a binary classification task, where each claim $x$ is assigned a label $y \in \{0,1\}$ indicating its correctness, following \citet{vazhentsev2026leveragingllmparametricknowledge}. Uncertainty scores serve as a proxy for correctness, with higher uncertainty interpreted as a higher probability of an incorrect claim. The benchmark is comprised of 9 datasets that cover diverse claim sources: human-written, rule-generated, short-form LLM generations, and excerpts parsed from long-form LLM outputs. 
On average, claims are 10--25 tokens long; exact per-dataset average lengths are shown in Figure~\ref{fig:weighting}. During evolutionary search, we use training subset of PopQA as dataset $D$ for candidate scoring. The performance is measured by AUC-ROC.

\paragraph{Selective prediction.}
We evaluate selective prediction by assigning each input $x$ a confidence score $s(x) \in \mathbb{R}$, which is used to rank predictions and compute performance under varying rejection thresholds. Evaluation is conducted on 8 datasets from \citet{fadeeva2023lmpolygraph}, covering tasks such as math reasoning, machine translation, and summarization. Importantly, this benchmark includes datasets with both short- and long-form answers. 

For evaluation, we first obtain model generations for each sample in each dataset, again using Llama-3.1-8B-Instruct. For comparison, average length of generations reaches 140 tokens for GSM8K, 68 tokens for SamSum, and 51 tokens for TruthfulQA dataset.
We then compute ground truth quality scores for each generation: ROUGE-L \citep{lin-2004-rouge} for SamSum, and TruthfulQA; exact match for BabiQA, MMLU and GSM8K; BLEU \citep{papineni-etal-2002-bleu} for WMT14 and WMT19; and F1 score for CoQA.


Performance is measured using the Prediction-Rejection Ratio (PRR). Compared to ROC-AUC, PRR naturally accommodates both discrete and continuous quality metrics, making it suitable for diverse generation tasks.
To compute PRR, first the rejection curves are built based on uncertainty estimates and ground truth quality of model predictions. Rejection curves represent average response quality as we abstain from the most uncertain predictions. PRR then is the ratio of the area between the UQ method and random baseline rejection curves to the area between the ideal and random baseline curves.  We compute PRR over the first 50\% of the curve, as higher rejection rates are typically impractical. The metric is normalized: PRR = 0 indicates random chance performance, while values near 1 indicate optimal performance.

\section{Results}

\subsection{Claim Verification}

We apply evolutionary search to design UQ methods for atomic claim verification. We conduct 20 runs using Gpt-oss-120B and 10 runs using Claude models, varying hyperparameters such as evolution prompt, LLM temperature, and number of candidates shown at each iteration to broadly explore the design space. For Claude-generated candidates, we select the top-3 candidates per run by validation performance on PopQA, yielding a shortlist of 30 methods. For Gpt-oss, this strategy produced many near-duplicate methods differing only in minor implementation details; we therefore manually curated a diverse subset of 16 methods among top scorers on the PopQA validation set. We then evaluate generalization on the test sets of PopQA and 8 additional atomic claim verification datasets from \citet{vazhentsev2026leveragingllmparametricknowledge}. The resulting average performance is shown in Figure~\ref{fig:barplot}.

\paragraph{Autonomously designed UQ methods surpass strong baselines in claim verification.} Both Gpt-oss-120B- and Claude-evolved methods outperform existing baselines by average ROC-AUC across all 9 datasets. Notably, only the PopQA training set is used for fitness evaluation; the remaining 8 datasets serve as out-of-distribution tests. Nevertheless, the evolved methods exhibit strong generalization. To assess statistical significance, we run a pairwise bootstrap difference test between all evolved methods and the strongest baseline, Sequence Probability. The results are given in Figure~\ref{fig:wintieloss}: the best evolved methods win on six and tie on one dataset.

\begin{table*}
\small
\begin{tabular}{llllllllllll}
\toprule
 & \makecell{Avg\\PRR} & BaBI & CoQA & MMLU & \makecell{Truth\\fulQA} & GSM8K & \makecell{WMT\\14 de} & \makecell{WMT\\14 fr} & \makecell{WMT\\19 de} & \makecell{WMT\\19 ru} & \makecell{SAM\\Sum} \\
\midrule
RAUQ & \textbf{54.6} & \textbf{79.2} & 80.2 & \textbf{75.9} & 37.2 & 88.5 & \underline{34.7} & 38.8 & 45.8 & 33.4 & 32.4 \\
\textcolor{teal}{SP$\times$ExpectedRank} & \textbf{54.6} & 78.7 & \textbf{80.6} & \textbf{75.9} & 39.1 & \underline{90.2} & 34.0 & 37.6 & 44.7 & 32.7 & \textbf{32.8} \\
\textcolor{teal}{SP$\times$MeanEntropy} & \textbf{54.6} & 78.0 & 79.8 & \underline{75.8} & 39.0 & \underline{90.2} & 34.2 & 38.0 & 45.4 & 33.0 & \textbf{32.9} \\
\textcolor{orange}{Evo-Anth-30} & \underline{54.5} & 78.1 & 79.0 & \underline{75.8} & \underline{39.3} & 90.1 & 34.0 & 37.8 & 45.1 & 32.7 & \textbf{32.8} \\
\textcolor{blue}{Evo-Gpt-9} & 54.4 & \textbf{79.2} & \underline{80.5} & \underline{75.8} & \textbf{39.4} & 89.9 & 33.3 & 37.0 & 44.0 & 32.2 & \underline{32.7} \\
\textcolor{blue}{Evo-Gpt-7} & 54.4 & \underline{79.1} & \underline{80.5} & \underline{75.8} & \textbf{39.4} & 89.9 & 33.3 & 37.0 & 44.0 & 32.2 & \underline{32.7} \\
SequenceProbability & 54.4 & \underline{79.1} & \underline{80.5} & \underline{75.8} & \textbf{39.4} & 89.9 & 33.3 & 37.0 & 44.0 & 32.2 & \underline{32.7} \\
\textcolor{orange}{Evo-Anth-1} & 54.3 & 78.2 & 77.5 & \underline{75.8} & 39.1 & \textbf{90.4} & 34.0 & 37.9 & 44.9 & 32.7 & \underline{32.7} \\
\textcolor{orange}{Evo-Anth-8} & 54.0 & 75.0 & 76.9 & \underline{75.8} & 39.1 & \textbf{90.4} & 34.2 & 38.1 & 45.0 & 32.8 & \underline{32.7} \\
\textcolor{orange}{Evo-Anth-22} & 54.0 & 76.0 & 79.4 & \underline{75.8} & \underline{39.3} & \textbf{90.3} & 33.2 & 37.1 & 44.0 & 32.2 & \underline{32.7} \\
Perplexity & 54.0 & 77.0 & 76.3 & \textbf{75.9} & 36.2 & 88.4 & \underline{34.7} & \underline{39.0} & \underline{46.1} & \underline{33.6} & 32.4 \\
\textcolor{orange}{Evo-Anth-15} & 53.7 & 74.6 & 79.1 & \underline{75.8} & \underline{39.3} & 90.1 & 33.1 & 36.9 & 43.8 & 32.1 & \underline{32.7} \\
MeanTokenEntropy & 53.7 & 73.7 & 76.7 & 75.1 & 36.1 & 88.4 & \textbf{34.8} & \textbf{39.1} & \textbf{46.5} & \textbf{33.8} & 32.6 \\
\textcolor{blue}{Evo-Gpt-5} & 52.5 & 67.7 & 73.3 & \underline{75.8} & 35.8 & 89.1 & 34.0 & 38.4 & 45.4 & 33.2 & 32.2 \\
\textcolor{blue}{Evo-Gpt-1} & 52.5 & 67.7 & 73.1 & \underline{75.8} & 35.8 & 89.0 & 34.1 & 38.5 & 45.4 & 33.3 & 32.2 \\
\textcolor{blue}{Evo-Gpt-13} & 52.1 & 69.2 & 75.8 & 72.7 & 35.4 & 89.8 & 33.1 & 37.2 & 43.7 & 32.4 & 31.8 \\
\textcolor{blue}{Evo-Gpt-11} & 51.0 & 72.0 & 72.2 & 70.1 & 34.1 & 86.7 & 31.9 & 36.7 & 42.9 & 31.4 & 31.6 \\
AttentionScore & 50.8 & 72.8 & 72.8 & 66.4 & \textbf{39.4} & 87.5 & 30.9 & 35.1 & 40.7 & 30.5 & 32.0 \\
\textcolor{blue}{Evo-Gpt-15} & 50.6 & 72.0 & 71.6 & 70.2 & 33.7 & 85.8 & 31.6 & 36.4 & 42.6 & 30.8 & 31.6 \\
\textcolor{blue}{Evo-Gpt-3} & 50.4 & 70.9 & 68.5 & 73.3 & 33.5 & 86.0 & 31.3 & 35.7 & 41.9 & 30.8 & 31.5 \\
\bottomrule
\end{tabular}
\caption{PRR scores across different methods and datasets (sorted by average rank). \textcolor{blue}{Blue} and \textcolor{teal}{teal}: Gpt-generated methods, \textcolor{orange}{orange}: Claude-generated methods. \textcolor{teal}{Teal} methods are evolved specifically for selective prediction, using training subsets of CoQA and SAMSum.}
\label{tab:prr_renamed}
\end{table*}

\subsection{Selective Prediction}

Selective prediction presents a more challenging test of generalization for two reasons. First, methods must capture fine-grained quality distinctions (e.g., between summaries with different ROUGE-L scores) rather than a binary correct/incorrect signal. Second, the substantially longer inputs may induce distribution shift relative to the short atomic claims used during evolution. We evaluate the evolved claim-verification methods in a zero-shot transfer setting on the selective prediction benchmark and, as a complementary experiment, launch additional evolution runs using Gpt-oss-120B with training splits of CoQA and SAMSum for candidate scoring. Table~\ref{tab:prr_renamed} reports PRR values for all estimators, ranked by average performance.

\paragraph{Cross-task transfer is limited, but in-domain evolution remains competitive.} Many methods evolved for claim verification underperform established baselines such as Sequence Probability and MeanTokenEntropy. This suggests that the binary verification objective does not fully capture the richer quality signal required for selective prediction. Interestingly, evolved methods are slightly ahead specifically on GSM8K, which has the longest average generations~--- despite being optimized for short atomic claims. In-domain evolution (\textcolor{teal}{teal}) yields methods that slightly exceed or match existing alternatives on average. Notably, these methods don't require access to attention maps, which is a limitation of RAUQ approach.

\section{Analysis}

We structure our analysis around several concrete research questions:

\begin{itemize}[noitemsep, leftmargin=25pt,topsep=3pt]
    \item[\textbf{RQ1:}]  How do search strategies differ across LLMs used in the evolutionary pipeline?
    \item[\textbf{RQ2:}] What recurring patterns emerge in autonomously designed methods?
    \item[\textbf{RQ3:}] Which input features are most useful for the evolutionary search?
    \item[\textbf{RQ4:}] How does method complexity relate to generalization performance across ?
    \item[\textbf{RQ5:}] How similar are evolved methods to the strongest baseline?
\end{itemize}

\subsection{RQ1: Search Strategy Differences}

To impose a simplicity prior, we include a free-form constraint in the evolution prompt: \textit{``Use up to 3 features.''}\ Gpt-oss-120B adheres to this instruction and consistently produces solutions that rely on at most three conceptually distinct features. Claude models, by contrast, systematically violate the constraint, generating significantly more complex methods that incorporate 10--30 features. Despite this, Claude-evolved methods ultimately achieve stronger generalization.
\begin{figure*}[t]
    \centering
    \includegraphics[width=\linewidth]{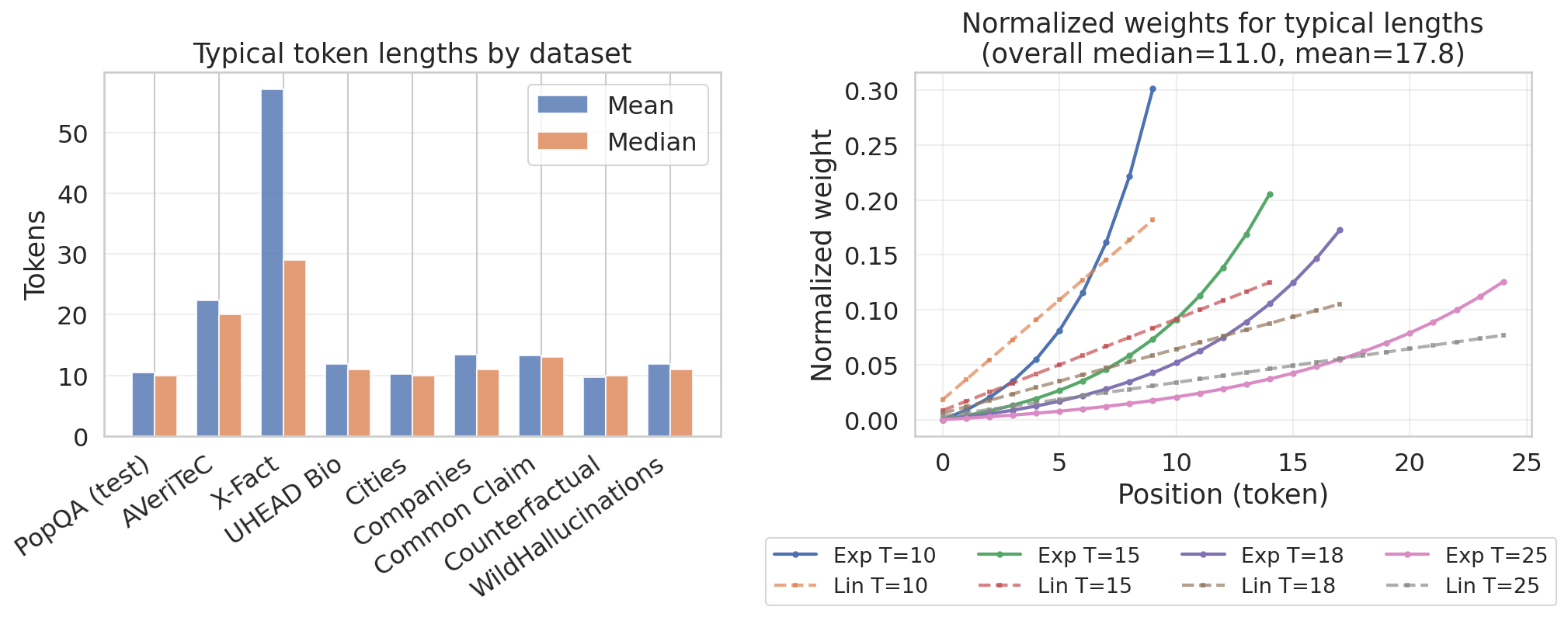}
    \caption{Left: average and median token lengths of atomic claims. Right: visualization of exponential and linear weighting for several typical claim lengths. Exponential weighting puts more emphasis on last tokens than linear weighting, especially for shorter sequences.}
    \label{fig:weighting}
\end{figure*}

This finding is noteworthy: classical statistical learning theory~\citep{Vapnik1998} predicts overfitting under high feature counts, particularly in the absence of explicit regularization. Moreover, the mechanisms commonly used to explain generalization in overparameterized deep networks such as implicit bias of SGD~\citep{liu2020understandingwhyneuralnetworksgeneralize,galanti2022sgdnoise} and loss landscape geometry~\citep{chiang2023losslandscapes} do not apply here, since evolutionary search constructs explicit programs without gradient-based optimization. We conjecture that the combination of the LLM's prior knowledge and the discrete nature of evolutionary selection acts as an implicit regularizer. This may represent a qualitatively distinct generalization mechanism specific to LLM-driven program synthesis.

\subsection{RQ2: Patterns in Evolved Methods}

We observe that positional information is a recurring motif in methods optimized for atomic claim verification. Among the estimators discovered by Gpt-oss-120B, position is leveraged in several distinct forms. Namely, Evo-Gpt-1 and Evo-Gpt-2 apply exponential weighting to token-wise log-probabilities, assigning greater importance to later positions in the sequence. The resulting weighting schemes are illustrated in Figure~\ref{fig:weighting}. Meanwhile, Evo-Gpt-3 computes the negated correlation between log-probabilities and token positions: sequences in which log-probabilities decline over the course of generation are scored as more uncertain.

\paragraph{Claude finds weights distinct from gradient descent.} The best-performing method developed by Claude is essentially a linear model with 36 custom features, without a single dominant signal. Thus, evolutionary search simultaneously performed feature engineering and weight tuning. A natural question is whether these weights are similar to what is found by training a logistic regression, and whether the found method can be further improved by choosing optimal weights.
To answer this, we trained two logistic regressions, based on feature sets of Evo-Anth-1 and Evo-Anth-15 candidates. 

\begin{table}[ht]
    \small
    \centering
    \begin{tabular}{lcc}
        \toprule
         & \textbf{AUC} & \textbf{95\% CI} \\
        \midrule
        Sequence Probability & 74.1 & (72.7, 75.4) \\
        Evolution (attention maps) & 75.1 & (73.8, 76.7) \\
        Evolution (logits) & 78.3 & (76.9, 79.6) \\
        Evolution (hidden states) & 74.1 & (72.7, 75.4) \\
        \midrule
        Evolution, T = 0.7 & 78.3 & (76.9, 79.6) \\
        Evolution, T = 1.0 & 78.7 & (77.4, 80.0) \\
        \makecell[l]{Evolution, T = 1.0 + \\domain knowledge} & 76.8 & (75.5, 78.0) \\
        \bottomrule
    \end{tabular}
    \caption{Ablation study, PopQA test split (candidates chosen by validation performance).}
    \label{tab:ablation}
\end{table}

Since we initially selected top-30 candidates by validation performance, Evo-Anth-15 should be representative of an honest median method with good validation but unknown test performance.

\begin{table*}[]
    \small
    \centering
    \begin{tabular}{lcccccccccc}
\toprule
 & \makecell{PopQA\\(train)} & \makecell{PopQA\\(test)} & AVeriTeC & X-Fact & UHead & Cities & Companies & \makecell{Common\\Claim} & \makecell{Counter\\fact.} & \makecell{Wild\\Hall.} \\
\midrule
Evo-1 & $0.84$ & $0.83$ & $0.68$ & $0.53$ & $0.68$ & $0.99$ & $0.89$ & $0.69$ & $0.78$ & $0.65$ \\
LogReg-1 & $\underset{\footnotesize \textcolor{DarkGreen}{+0.02}}{0.86}$ & $\underset{\footnotesize \textcolor{DarkGreen}{+0.01}}{0.84}$ & $\underset{\footnotesize \textcolor{red}{-0.04}}{0.65}$ & $\underset{\footnotesize \textcolor{DarkGreen}{+0.04}}{0.57}$ & $\underset{\footnotesize \textcolor{DarkGreen}{+0.01}}{0.69}$ & $\underset{\footnotesize \textcolor{red}{-0.01}}{0.98}$ & $\underset{\footnotesize \textcolor{red}{-0.02}}{0.87}$ & $\underset{\footnotesize \textcolor{red}{-0.03}}{0.67}$ & $\underset{\footnotesize \textcolor{red}{-0.01}}{0.77}$ & $\underset{\footnotesize \textcolor{gray}{0.00}}{0.65}$ \\
\midrule
Evo-15 & 0.82 & 0.81 & 0.68 & 0.51 & 0.67 & 0.99 & 0.84 & 0.67 & 0.76 & 0.66 \\
LogReg-15 & $\underset{\footnotesize \textcolor{DarkGreen}{+0.03}}{0.85}$ & $\underset{\footnotesize \textcolor{DarkGreen}{+0.02}}{0.83}$ & $\underset{\footnotesize \textcolor{red}{-0.03}}{0.65}$ & $\underset{\footnotesize \textcolor{gray}{0.00}}{0.51}$ & $\underset{\footnotesize \textcolor{DarkGreen}{+0.02}}{0.68}$ & $\underset{\footnotesize \textcolor{red}{-0.01}}{0.99}$ & $\underset{\footnotesize \textcolor{DarkGreen}{+0.06}}{0.90}$ & $\underset{\footnotesize \textcolor{DarkGreen}{+0.01}}{0.68}$ & $\underset{\footnotesize \textcolor{DarkGreen}{+0.03}}{0.80}$ & $\underset{\footnotesize \textcolor{red}{-0.02}}{0.63}$ \\
\bottomrule
\end{tabular}
    \caption{Results of training logistic regressions on  feature sets of Evo-Anth-1 and Evo-Anth-15 canditates. LogReg-X corresponds to a logistic regression employing Evo-X features. Notable, weights found via evolutionary search generalize slightly better to OOD datasets, despite both approaches only used PopQA train set. Displayed deltas sometimes disagree with AUC values due to rounding; in these cases deltas are more precise.}
    \label{tab:claude_log_reg}
\end{table*}

Notably, logistic regression finds very different sets of weights compared to evolution. Correlation between vectors of coefficients found by evolution and logistic regression equals $-0.23$ for Evo-Anth-1, and $0.06$ for Evo-Anth-15, which suggests that the two methods occupy qualitatively different basins in weight space. Nonetheless, the performance is comparable, as shown in Table \ref{tab:claude_log_reg}.

\subsection{RQ3: Utility of Input Features}

We consider three categories of input features available from the target LLM: logits, attention maps, and hidden states from the residual stream. To determine which feature types are most beneficial in this setting, we conduct a controlled ablation study in which only a designated subset of features is made available to evolved candidates. All configurations retain access to basic logit-derived statistics (per-token log-probabilities and entropies), as methods relying solely on attention or hidden-state features yielded near-random performance. In the \texttt{logits}, \texttt{attention}, and \texttt{hidden states} regimes, we additionally provide the top-512 logits per token, full attention maps, or embeddings from all tokens and layers, respectively. All runs use Gpt-oss-120B as the evolution LLM and report test ROC-AUC on PopQA.

As shown in Table \ref{tab:ablation}, access to attention maps and hidden-state features does not improve performance beyond the logits-only condition. Manual inspection of generated candidates reveals that evolution frequently converges on the same rudimentary derived features~--- such as the entropy of attention distributions or the average Euclidean distance between consecutive token embeddings~--- none of which provide meaningful uplift.

We additionally investigate the effects of sampling temperature and domain knowledge injection. Higher temperature yields modestly higher top scores, likely due to increased candidate diversity. Incorporating domain knowledge into the evolution prompt, however, does not improve results (Table~\ref{tab:ablation}, lower half).

\subsection{RQ4: Method Complexity and Generalization}

\begin{figure}
    \centering
    \includegraphics[width=\linewidth]{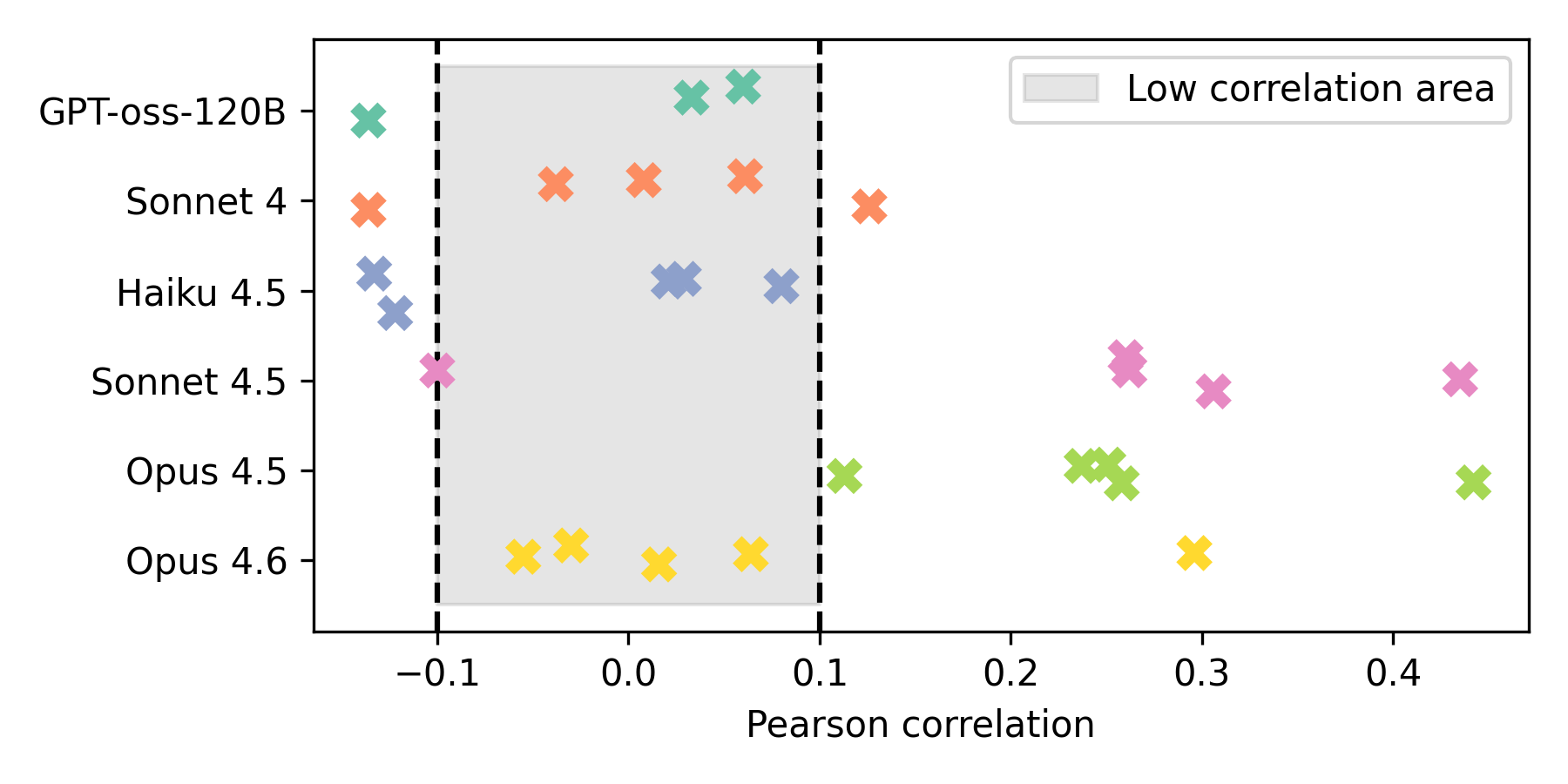}
    \caption{Correlation between method complexity and performance for different LLMs. Each dot corresponds to one evolution run.}
\label{fig:complexity_performance_correlation}
\end{figure}

A natural question is whether LLMs can design complex UQ methods without overfitting, and whether this capability scales with model size. To investigate, we measure the Spearman correlation between method complexity (operationalized as line count) and test ROC-AUC on PopQA for each evolution LLM. As shown in Figure~\ref{fig:complexity_performance_correlation}, only Sonnet~4.5 and Opus~4.5 exhibit a consistent positive relationship between complexity and performance. Surprisingly, Opus~4.6 shows a regression relative to Opus~4.5, suggesting that the relationship between model capability and effective use of complexity is non-monotonic. We conjecture this may reflect reduced candidate diversity in Opus 4.6's outputs, though this remains untested.

\begin{figure*}
    \centering
    \includegraphics[width=\linewidth]{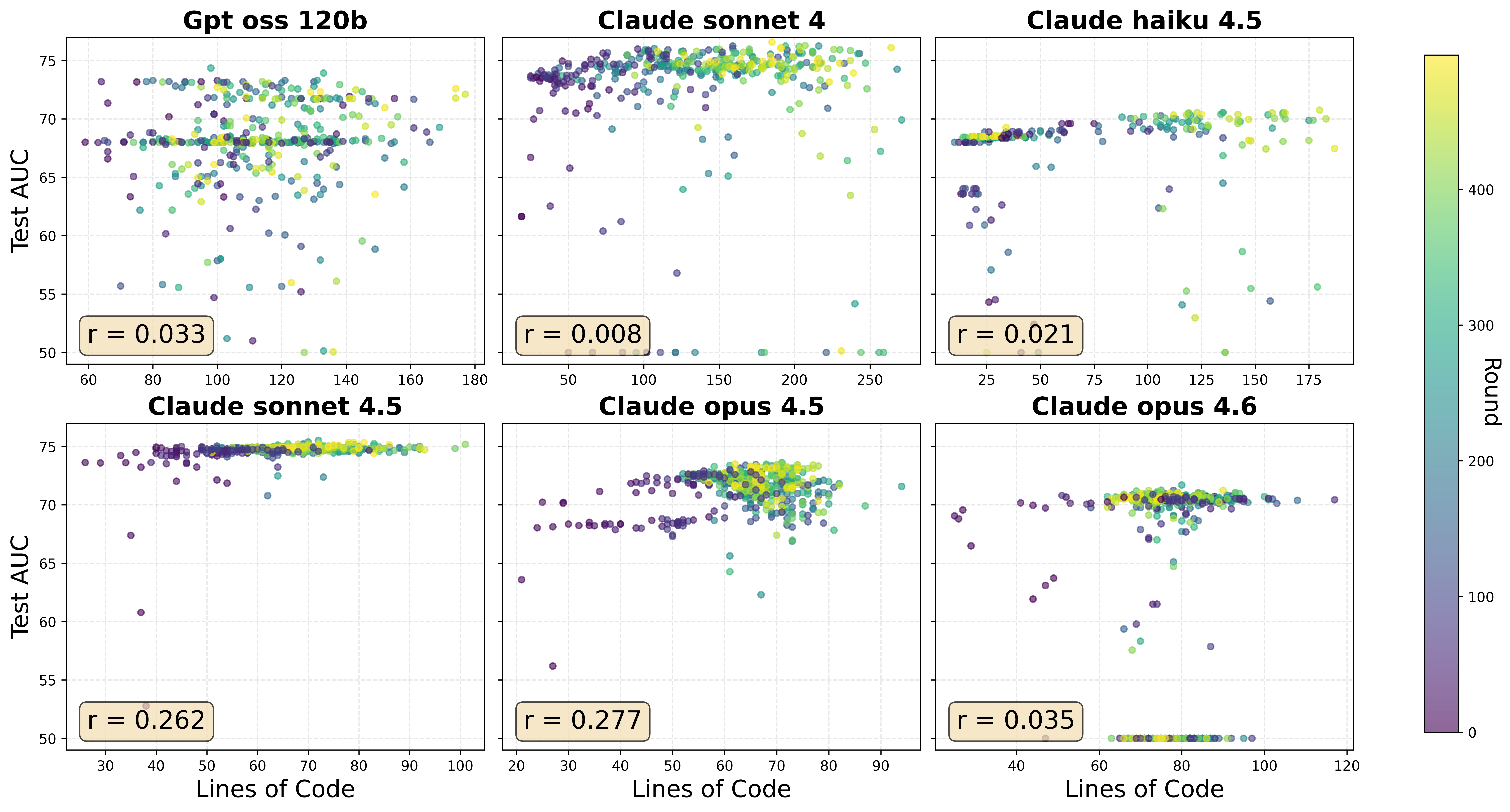}
    \caption{Evolution dynamics in (complexity, performance) coordinates for 6 different models. We use line count as the simplest proxy for method complexity; other proxies such as amount of AST nodes, number of binary and unary operators, or Halstead volume follow the same pattern.}
    \label{fig:evolution_cloud}
\end{figure*}

To verify that this finding is not driven by outliers, we visualize the full evolution trajectory~--- 500 rounds of candidates plotted in (complexity, performance) space~--- for one representative run per model (Figure~\ref{fig:evolution_cloud}). Opus~4.6 indeed shows no upward trend: candidates cluster around two performance levels (${\sim}70$ and ${\sim}50$ ROC-AUC) irrespective of complexity. Haiku~4.5 and Sonnet~4.0 exhibit a slight upward drift that is masked by outliers in the aggregate correlation statistic. Gpt-oss-120B produces a notably more dispersed candidate distribution, reflecting its broader but less targeted exploration of the design space.

\subsection{RQ5: Similarity to Existing Baselines}
\label{sec:similarity_to_baseline}

Sequence Probability is the strongest baseline for atomic claim verification, and many evolved methods build upon it as a base signal. This raises a natural question: are the evolved methods fundamentally different from Sequence Probability, or do they constitute minor variations of the same underlying approach?

To answer this, we introduce a data-driven notion of UQ method similarity. For each method, we compute its predictions on a dataset of $N$ atomic claims, yielding a fixed-length score vector. We then measure Spearman correlation between these vectors for any two methods. This approach captures the degree to which two methods produce similar rankings over inputs, regardless of differences in scale or functional form.

To construct a representative evaluation set, we sample examples from PopQA, X-Fact, and UHead training subsets in equal proportions, yielding a mixture of 5{,}536 claims. We then compute the similarity between Sequence Probability and all candidate methods from a single Gpt-oss-120B evolution run, alongside their test performance (Figure~\ref{fig:spearman_candidates}).
Notably, several evolved methods exhibit near-zero Spearman correlation with Sequence Probability, indicating that they produce qualitatively different uncertainty rankings while achieving comparable performance. This confirms that evolutionary search can discover diverse UQ approaches.

\section{Conclusion}
We applied LLM-powered evolutionary search to the automated design of UQ methods and evaluated the resulting methods across 17 datasets spanning two tasks. On atomic claim verification, evolved methods surpass strong existing baselines, achieving up to 6.7\% relative ROC-AUC improvement across 9 datasets and generalizing robustly out-of-distribution despite using only a single training dataset for candidate scoring. On selective classification, direct transfer of claim-verification methods is limited; however, in-domain evolution recovers competitive performance, suggesting the framework is adaptable rather than task-specific.

Qualitative analysis reveals two distinct search strategies. Gpt-oss-120B produces interpretable, low-complexity estimators that rely heavily on token positions. Claude models instead construct high-dimensional linear estimators with 10–30 features that, unexpectedly, generalize well out-of-distribution~--- and whose weights are largely uncorrelated with those found by logistic regression, pointing to a qualitatively different solution regime than gradient descent. Analysis of evolution dynamics shows that the ability to leverage complexity effectively is model-dependent: Sonnet 4.5 and Opus 4.5 exhibit a consistent positive complexity-performance relationship, while Gpt-oss-120B, Sonnet 4, Haiku 4.5, and — surprisingly — Opus 4.6 do not. We conjecture that the latter may reflect reduced output diversity in Opus 4.6, though this remains to be verified.

Beyond hallucination detection, these results suggest that frontier LLMs differ not just in capability but in how they explore open-ended search spaces. This deserves further attention since LLM-powered optimization is applied to increasingly complex scientific and engineering problems.

\section*{Limitations}

While the results are encouraging, several limitations should be kept in mind. Our evolutionary search is performed using feedback from a single training dataset, PopQA. Although the resulting methods generalize well across the atomic claim verification benchmark, this setup still leaves open the possibility that some useful design choices are tied to properties of that dataset.

The study also focuses on a deliberately constrained search space: we optimize unsupervised, lightweight, and interpretable Python programs built over precomputed model statistics. This makes the discovered methods easy to analyze and inexpensive at inference time, but it also means that more expressive detector families, including supervised or neural approaches, are outside the scope of the current work.

In addition, our experiments rely on internal signals extracted from a single generator model, Llama-3.1-8B-Instruct. As a result, the extent to which the evolved estimators transfer across model families, scales, or decoding regimes remains an open question. A related point is that our evaluation covers two settings, atomic claim verification and selective prediction, which already provide a meaningful range of tasks, but do not capture the full diversity of hallucination phenomena encountered in long-form generation, interactive dialogue, or tool-augmented systems.

Finally, several of our observations are primarily empirical. In particular, the strong generalization of some higher-dimensional evolved estimators, as well as the differences in search behavior across LLMs, are interesting findings, but their underlying mechanisms are not yet fully understood.


\bibliography{custom}

\appendix



\section{Additional plots for RQ5: Similarity to Existing Baselines}

\begin{figure}
    \centering
    \includegraphics[width=\linewidth]{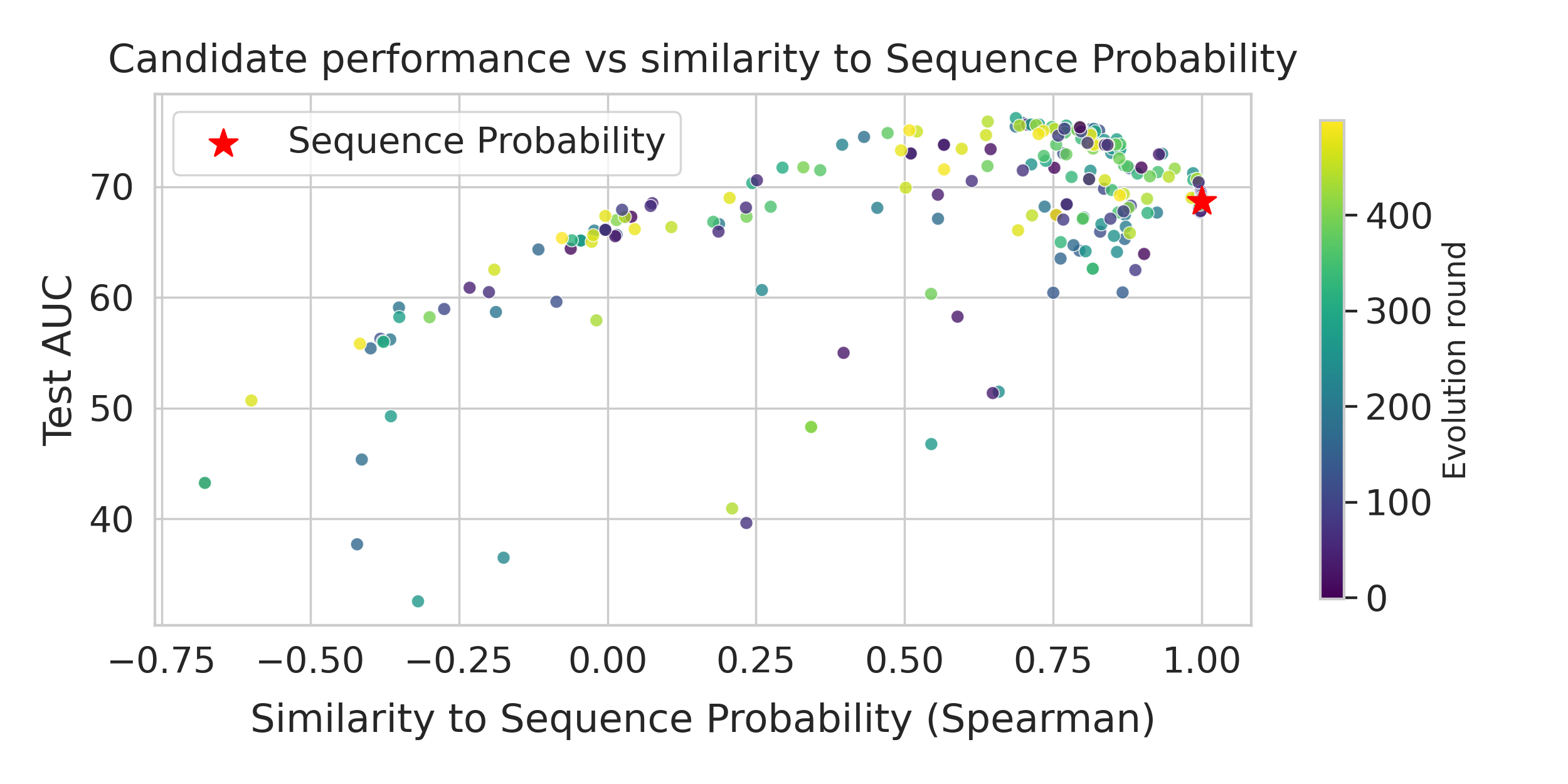}
    \caption{Methods' similarity to SequenceProbability (measured as Spearman correlation between predictions of method and SequenceProbability) versus test performance. Notably, there exist methods which have accuracy comparable to SequenceProbability, but producing significantly different sample rankings (with Spearman correlation even less than 0).}
    \label{fig:spearman_candidates}
\end{figure}

Sequence Probability is used as a starting point for most of the experiments throughout the paper. How similar are the evolved candidates to their starting point, and how this similarity changes over the course of evolution?  To answer this question, we measure similarity between Sequence Probability and all candidates from a selected evolution run, and plot it on Figure \ref{fig:spearman_candidates}.
The methodology for similarity estimation is detailed in Section \ref{sec:similarity_to_baseline}.

\section{Statistical significance testing for atomic claim verification results}
\begin{figure}[h]
    \centering
    \includegraphics[width=\linewidth]{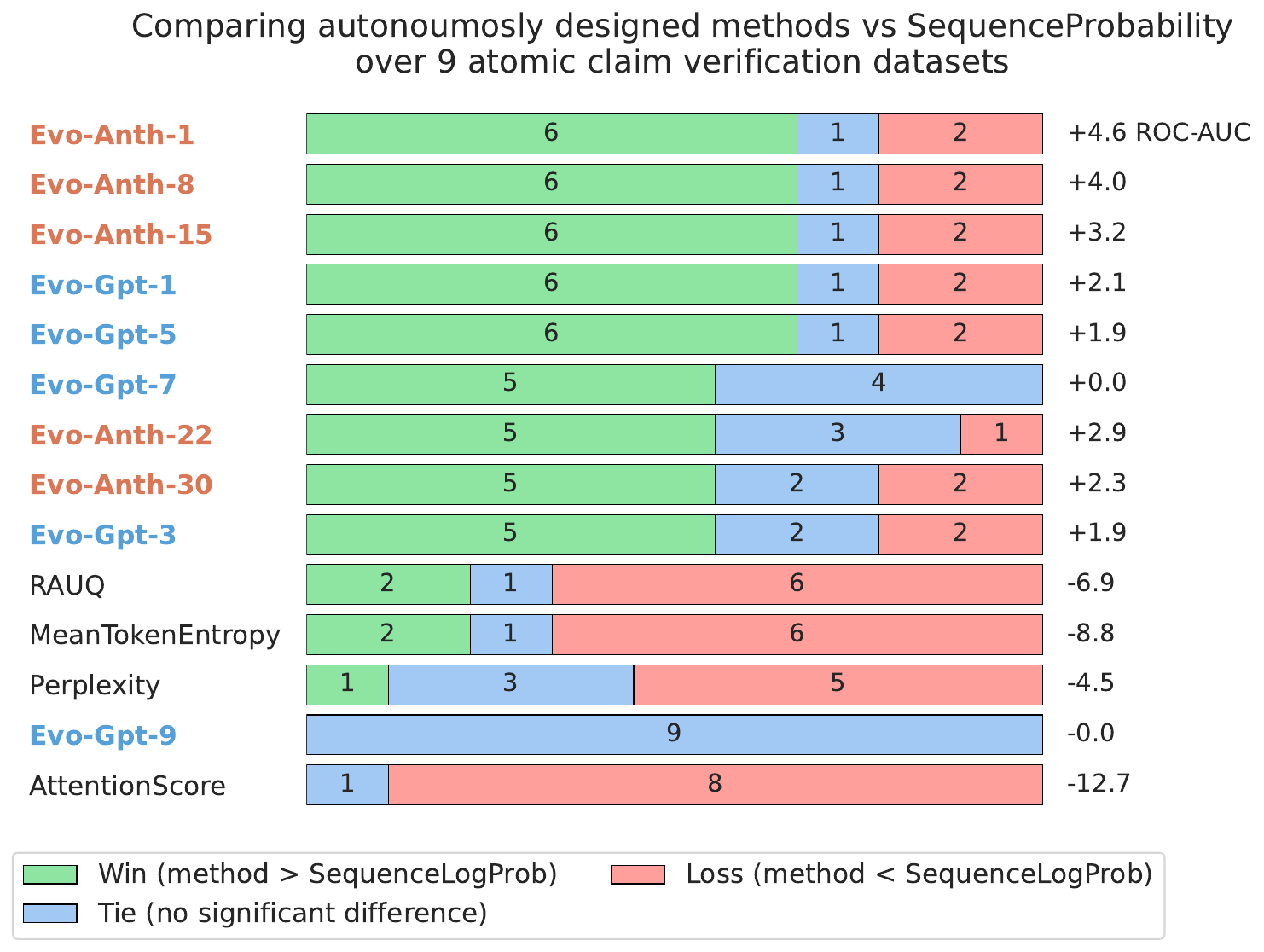}
    \caption{Best autonomously designed methods deliver statistically significant improvement over the strongest baseline (SequenceLogProb) on 6/9 datasets, and tie on one. We use paired bootstrap difference test with Bonferroni correction. Orange methods are designed using Claude models, and blue methods are designed using Gpt-oss-120B.}
    \label{fig:wintieloss}
\end{figure}

To test whether the improvements of evolved uncertainty quantification methods are in ROC-AUC on atomic claim verification datasets are statistically significant, we run paired bootstrap difference test between Sequence Probability (the strongest baseline) and all other methods. We run comparisons on each dataset separately. Figure \ref{fig:wintieloss} shows aggregated results. Best evolved methods provide significant improvements on 6 datasets, and tie on 1.

\end{document}